\documentclass[10pt,twocolumn,letterpaper]{article}
\usepackage{cvpr}
\usepackage{times}
\usepackage{epsfig}
\usepackage{graphicx}
\usepackage{amsmath}
\usepackage{amssymb}
\usepackage{tabularx}
\usepackage{authblk}
\usepackage[usenames, dvipsnames]{color}
\usepackage{caption}
\usepackage[pagebackref=true,breaklinks=true,letterpaper=true,colorlinks,bookmarks=false]{hyperref}
\cvprfinalcopy 
\def\cvprPaperID{1708} 

\ifcvprfinal\fi
\begin{document}
\title{Detailed, accurate, human shape estimation from clothed 3D scan sequences}

\author[1,2]{Chao Zhang}
\author[1]{Sergi Pujades}
\author[1]{Michael Black}
\author[1]{Gerard Pons-Moll}

\affil[1]{MPI for Intelligent Systems, T\"ubingen, Germany}
\affil[2]{Dept. of Computer Science, The University of York, UK}

\twocolumn[{%
\renewcommand\twocolumn[1][]{#1}%
\maketitle
\begin{center}
    \newcommand{\teaserwidth}{1.\textwidth}
\vspace{-0.4in}
    \centerline{
    \includegraphics[width=\teaserwidth,clip]{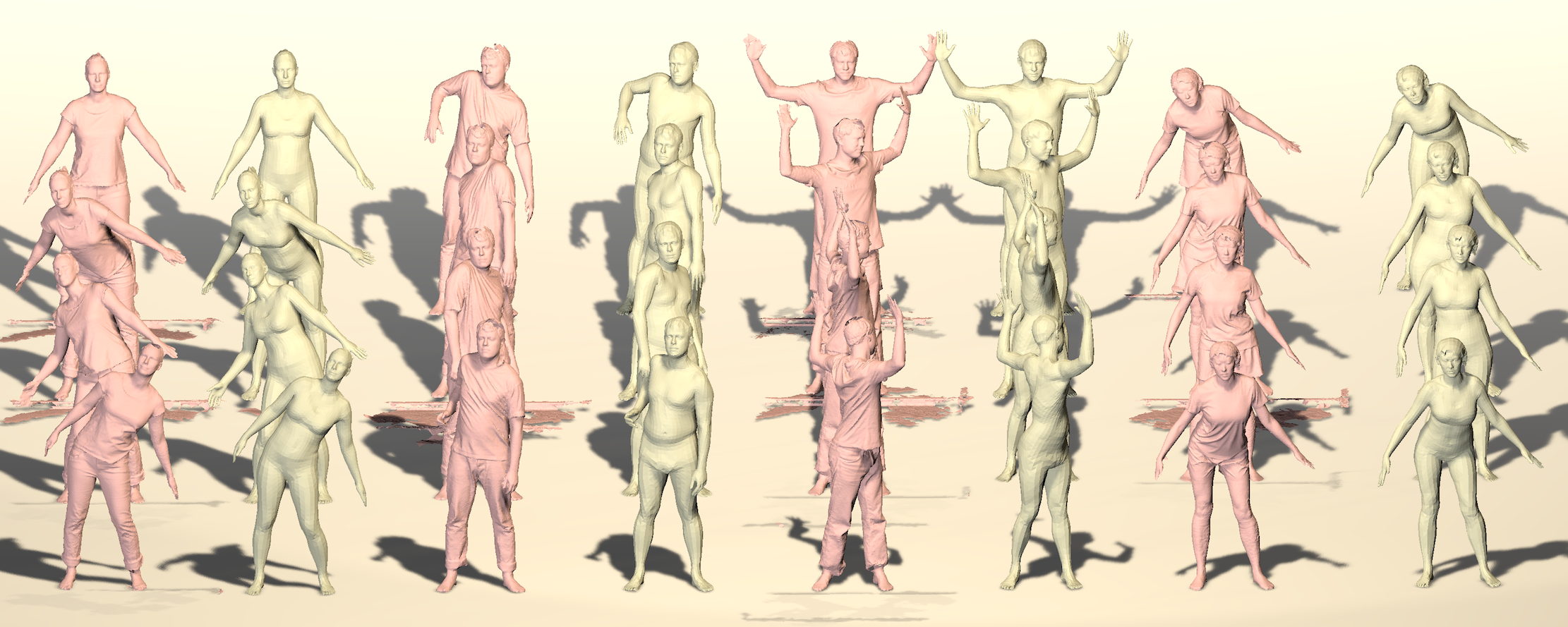}
    }
\vspace{-0.1in}
    \captionof{figure}{Given static 3D scans or 3D scan sequences (in pink),
    we estimate the naked shape under clothing (beige).  
    Our method obtains accurate results by minimizing an objective function
    that captures the visible details of the skin, while being robust to clothing.
    We show several pairs of clothed scan sequences and the estimated body shape underneath.
}
        \label{fig:teaser}
\end{center}%
}]
\maketitle
\thispagestyle{empty}

\begin{abstract}
\vspace{-0.16in}
We address the problem of estimating human pose and body shape from 3D scans over time. Reliable estimation of 3D body shape is necessary for many applications including virtual try-on, health monitoring, and avatar creation for virtual reality. Scanning bodies in minimal clothing, however, presents a practical barrier to these applications. We address this problem by estimating body shape under clothing from a sequence of 3D scans. Previous methods that have exploited body models produce smooth shapes lacking personalized details. We contribute a new approach to recover a personalized shape of the person. The estimated shape deviates from a parametric model to fit the 3D scans. We demonstrate the method using high quality 4D data as well as sequences of visual hulls extracted from multi-view images.  We also make available BUFF, a new 4D dataset that enables quantitative evaluation~{\small \url{http://buff.is.tue.mpg.de/}}. Our method outperforms the state of the art 
in both pose estimation and shape estimation,
qualitatively and quantitatively.
\end{abstract}

\newcommand{\dataset}{{BUFF }}

\newcommand{\vect}[1]{\mathbf{#1}}
\newcommand{\mat}[1]{\vect{#1}}
\newcommand{\matelem}[2]{\mat{#1}_{#2}}
\newcommand{\vecelem}[2]{\vect{#1}_{#2}}
\newcommand{\argmin}[1]{\underset{#1}{\operatorname{arg\,min}}}
\newcommand{\partref}[1]{Part~\ref{#1}}
\newcommand{\chapref}[1]{Chapter~\ref{#1}}
\newcommand{\eqnref}[1]{Eq.~(\ref{#1})}
\newcommand{\sectref}[1]{Sec.~\ref{#1}}
\newcommand{\figref}[1]{Fig.~\ref{#1}}
\newcommand{\tabref}[1]{Tab.~\ref{#1}}

\newcommand{\R}{\mathbb{R}}
\newcommand{\pose}{\boldsymbol{\theta}}                 
\newcommand{\poseGT}{\boldsymbol{\theta}_{GT}}          
\newcommand{\poseEst}{\boldsymbol{\theta}_{Est}}        
\newcommand{\trans}{\vect{t}}                     
\newcommand{\shape}{\boldsymbol{\beta}}                 
\newcommand{\nakedshape}{\mat{T}_n}              
\newcommand{\vt}{\mat{T}_\mathrm{\mu}}                        
\newcommand{\vtGT}{\mat{T}_\mathrm{GT}}                 
\newcommand{\vtEst}{\mat{T}_\mathrm{Est}}			   
\newcommand{\vtCloth}{\mat{T}_\mathrm{cloth}}			   
\newcommand{\vtFusion}{\mat{T}_\mathrm{Fu}}

\newcommand{\scan}{\mathcal{S}} 
\newcommand{\fusionscan}{\mathcal{S_\mathrm{Fu}}}
\newcommand{\scanvertex}{\vect{s}} 

\newcommand{\algn}{\mathcal{A}}                      
\newcommand{\joints}{\mat{J}}
\newcommand{\posebs}{B_p(\pose)}          
\newcommand{\shapebs}{B_s(\shape)}          
\newcommand{\posebsvertex}{b_p(\pose)}          
\newcommand{\shapebsvertex}{b_s(\shape)}          
\newcommand{\vtvertex}{\vect{v}}   

\newcommand{\cloth}{\mathcal{S}_\mathrm{cloth}}             
\newcommand{\skin}{\mathcal{S}_\mathrm{skin}}               
\newcommand{\nframes}{N_\mathrm{frames}}           
\newcommand{\nverts}{N_\mathrm{verts}} 
\newcommand{\smpl}{M}                   
\newcommand{\smplsurf}{\mathcal{M}}                   
\newcommand{\bweights}{\mat{W}}
\newcommand{\SP}[1]{\textcolor{blue}{SP: #1}}
\newcommand{\CZ}[1]{\textcolor{red}{CZ: #1}}
\newcommand{\GP}[1]{\textcolor{cyan}{GP: #1}}
\newcommand{\red}[1]{\textcolor{red}{#1}}

\newcommand{\slabel}{v}
\newcommand{\slabels}{\vect{v}}
\newcommand{\mpoint}{\vect{x}}
\newcommand{\apoint}{\vect{x}}
\newcommand{\spoint}{\vect{x}}

\section{Introduction}
\label{sec:introduction}
We address the problem of estimating the body shape of a person
wearing clothing from 3D scan sequences or visual hulls computed from multi-view images. 
Reliably estimating the shape under clothing is useful for many applications including virtual try-on, biometrics, and fitness. 
It is also a key component for virtual clothing and cloth simulation
 where garments need to be synthesized on top of the minimally-clothed body. 
Furthermore, most digital recordings of humans are done wearing clothes and therefore
automatic methods to extract biometric information from such data are needed.
While clothes occlude the minimally-clothed shape (MCS) of the human and make the task challenging,
different poses of the person provide different constraints on the shape under the clothes.
Previous work \cite{bualan2008naked, yang2016estimation} exploits this
fact by optimizing shape using different poses.
They use the statistical shape model SCAPE~\cite{anguelov2005scape} that factorizes human shape into subject identity and pose.
The main limitation of such approaches is that only the parameters of the statistical model
are optimized and so the solutions are constrained to lie on the model space.
While statistical models provide powerful constraints on the human shape, 
they are typically
overly-smooth and important identity details such as face features are lost.
More importantly, constraints such as ``the cloth garment should lie
outside the body shape surface'' are difficult to satisfy when optimizing model parameters.
This is because shape deformations in most statistical body models are
global, so a step in model space that, for example, shrinks the belly might
have the ``side effect'' of making the person shorter. 
Therefore, we propose a novel method to estimate the MCS,
that recovers accurate global body shape as well as
important local shape identity details as can be seen in \figref{fig:teaser}. 
Our hypothesis is that several poses of a person wearing the same clothes
provide enough constraints for detailed body shape capture. 
Moreover, if identity details are visible, e.g.~the face, the method should capture them. 

To do so, we propose to minimize a single-frame objective function
that 
(i) enforces the scan cloth vertices to remain outside of the MCS, 
(ii) makes the MCS tightly fit the visible skin parts, and
(iii) uses a robust function that snaps MCS to close-by cloth vertices 
and ignores far away cloth points.

In contrast to previous work, where only model shape parameters are optimized,
we directly optimize the $N=6890$ vertices of a template in 
a canonical ``T'' pose (unposed template). 
This allows us to capture local shape details by satisfying the objective constraints.
To satisfy anthropometric constraints, we regularize the optimisation
vertices to remain close to a statistical body model.
We use SMPL \cite{loper2015smpl}, a publicly available vertex-based model that is 
compatible with standard graphics pipelines.
While this formulation has a larger number of variables to optimise,
we show that it leads to more accurate and more detailed results.

While simple, the proposed single-frame objective is powerful, 
as it can be adapted to different tasks. 
To leverage the temporal
information one would like to optimize all scans in the sequence
at once. 
However, given high resolution scans,
this is computationally very expensive and memory intensive.
%
Hence, we first register/align all scans by deforming one template to explain
both, skin and cloth scan points. These \emph{cloth alignments} are obtained by
minimizing a special case of the single-frame objective treating all scan vertices as skin.
Since the model factors pose and shape, all cloth alignment templates
live in a common unposed space; we call the union of these unposed alignments 
the {\em fusion scan}. 
Since the cloth should lie outside the body for all frames
we minimize the single-frame objective using the fusion scan as input
and obtain an accurate shape template (\emph{fusion shape}) for the person.
Finally, to obtain the pose and the time varying shape details, we optimize again 
the single objective function using the fusion shape as a regularizer.
The overview of the method is described in \figref{fig:outline}. The result is a 
numerically and visually accurate estimation of the body shape under clothing
and its pose that fits the clothed scans (see \figref{fig:teaser}).

To validate our approach we use an existing dataset \cite{yang2016estimation}
and collected a new dataset (\dataset: Bodies
Under Flowing Fashion)
that includes high resolution 3D scan sequences of 3 males and 3 females in different clothing styles.
We make \dataset publicly available for research 
purposes at \url{http://buff.is.tue.mpg.de/}. \dataset contains in total $11,054$ high resolution clothed scans with ground
truth naked shape for each subject. Qualitative as well as quantitative results demonstrate 
that our method outperforms previous state of the art methods.

\section{Related Work}
\label{sec:related_work}
{\bf Body Models.} 
A key ingredient for robust human pose and shape estimation is the body model~\cite{PonsModelBased}. Early body models in computer vision were based on simple primitives 
\cite{barron2000estimating, gavrila1996, plankers2001articulated, sigal2004tracking}. More recent body models~\cite{anguelov2005scape, loper2015smpl, zuffi2015stitched} 
encode shape and pose deformations separately and are learned from thousands of scans of real people. Some works model shape and pose deformations jointly as 
in \cite{hasler2009statistical} where they perform PCA on a rotation-invariant encoding of triangles.

A popular body model is SCAPE~\cite{anguelov2005scape}, which factors triangle deformations into pose and shape. Recent work has proposed to make SCAPE more
efficient by approximating the pose dependent deformations with Linear Blend Skinning (LBS)~\cite{jain2010,fscape}. 
To increase the expressiveness of the shape space,~\cite{chen2013tensor} combines SCAPE with localized multilinear models for each body part. 
SMPL~\cite{loper2015smpl} models variations due to pose and shape using a linear function. Some models~\cite{Park08,pons2015dyna} incorporate 
also dynamic soft-tissue deformations; inferring soft-tissue deformations under clothing is an interesting future direction.

{\bf Pose and Shape Estimation.} 
A large number of works estimate the body pose and shape from people wearing minimal clothing. 
Methods \cite{weiss2011home,ye2014real,zhang2014quality} to estimate pose and shape from a depth sensor typically combine silhouette, depth data or color terms. 
In \cite{Bogo:ICCV:2015} they estimate body shape from a depth sequence but they focus on people wearing minimal clothing. 

In \cite{Pons-Moll_MRFIJCV} they estimate the pose and shape from depth data combining bottom up correspondences with top down model fitting. However,
clothing is not explicitly modeled.
In \cite{helten2013personalization} they propose a real-time full body tracker based on Kinect but they first acquire the shape of the subject in a fixed posture 
and then keep shape parameters fixed.

A large number of methods track the human pose/shape from images or multi-view images ignoring clothing or treating it as noise 
\cite{balan2007detailed,gall2010optimization,stoll2011fast}. 
The advent of human part detectors trained from large amounts of annotated data using Convolutional Neural Networks 
\cite{deepcut, CPM, hourglass} has made human shape and pose estimation possible in challenging scenarios~\cite{BogoECCV2016, rhodin2016general,dibra2016shape,Lassner}.  
In \cite{BogoECCV2016} they fit a SMPL model to joint detections to estimate pose and shape.
However the estimated shape is a simplification since bone lengths alone 
can not determine the full body shape.
Recently, \cite{rhodin2016general} uses a sum-of-Gaussians body model~\cite{stoll2011fast} and estimates pose 
and shape in outdoor sequences but the alignment energy does not consider clothing.

{\bf Shape Under Clothing.}
Estimating the underlying shape occluded by cloth is a highly under-constrained problem.
To cope with this, most existing methods
exploit statistical body models, like SCAPE or variants of it. 
In \cite{hasler2009estimating}  they estimate shape from a single 3D scan. 
Their rotation-invariant body representation does not separate 
and pose parameters and thus it can not be trivially extended to sequences. In~\cite{neophytou2014layered} they propose a layered model of cloth 
and estimate the body shape by detecting areas where the cloth is close to the body. 
Wuhrer et al.~\cite{wuhrer2014estimation} estimate shape under clothing on single or multiple 3D scans.
The pose and shape is estimated at every frame and the final shape is obtained as the average over multiple frames. 
Stoll et al.~\cite{stoll2010video} estimate the naked shape under a clothed template but require manual input and their focus 
is on estimating an approximate shape to use as a proxy for collision detection. 
All these approaches
require manual input to initialize the pose~\cite{hasler2009estimating,neophytou2014layered,wuhrer2014estimation}.

The work of \cite{rosenhahn2007system} incorporates a model of clothing for more robust tracking under clothing 
but only results for the lower leg are shown and shape is given as input to the method.
Following the same principle~\cite{guan20102d}, proposes to learn the statistics of how cloth deviates from the body for robust inference but they do so in 2D. 
Similarly in \cite{song20163d} they dress the SCAPE body model using physics simulation to learn such statistics but the clothing variety is very limited. 

The authors of~\cite{bualan2008naked} estimate the body shape under clothing from multi-view images and like us they exploit temporal information.
However, they only optimize model parameters and hence shape details are not captured.  Numerical evaluation is only provided using biometric shape features.
The work of \cite{yang2016estimation} proposes a similar approach to estimate shape and pose under clothing in motion but they do it from scans and only optimize model parameters.  
The pose deformation model used in~\cite{yang2016estimation} is too simple to track complex poses such as shrugging or lifting arms.

The previous work is restricted to optimize model parameters and hence, the results lack detail because they are restricted to the model space.
We go beyond state of the art and estimate jointly model parameters and a subject specific free-form shape.
Other work is model free and estimates non-rigid 3D shape over time 
\cite{dou2016fusion4d,newcombe2015dynamicfusion,orts2016holoportation,zollhofer2014real}.  
While this work can capture people in clothing, it does not use a body model and cannot estimate shape {\em under} clothing.
Our method combines the strong constraints of a body model with the freedom to deform like the model-free methods.

\section{Body Model}
\label{sec:body_model}
\begin{figure*}[ht!]
\centering
\begin{tabular}{ccccc}
\includegraphics[height=2.8cm]{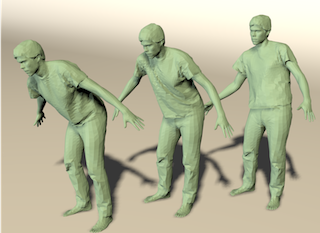}\llap{a~} &
\includegraphics[height=2.8cm]{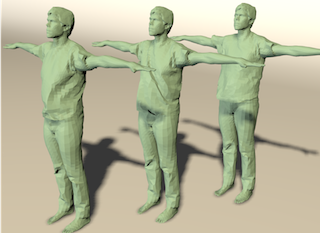}\llap{b~} & 

\includegraphics[height=2.8cm]{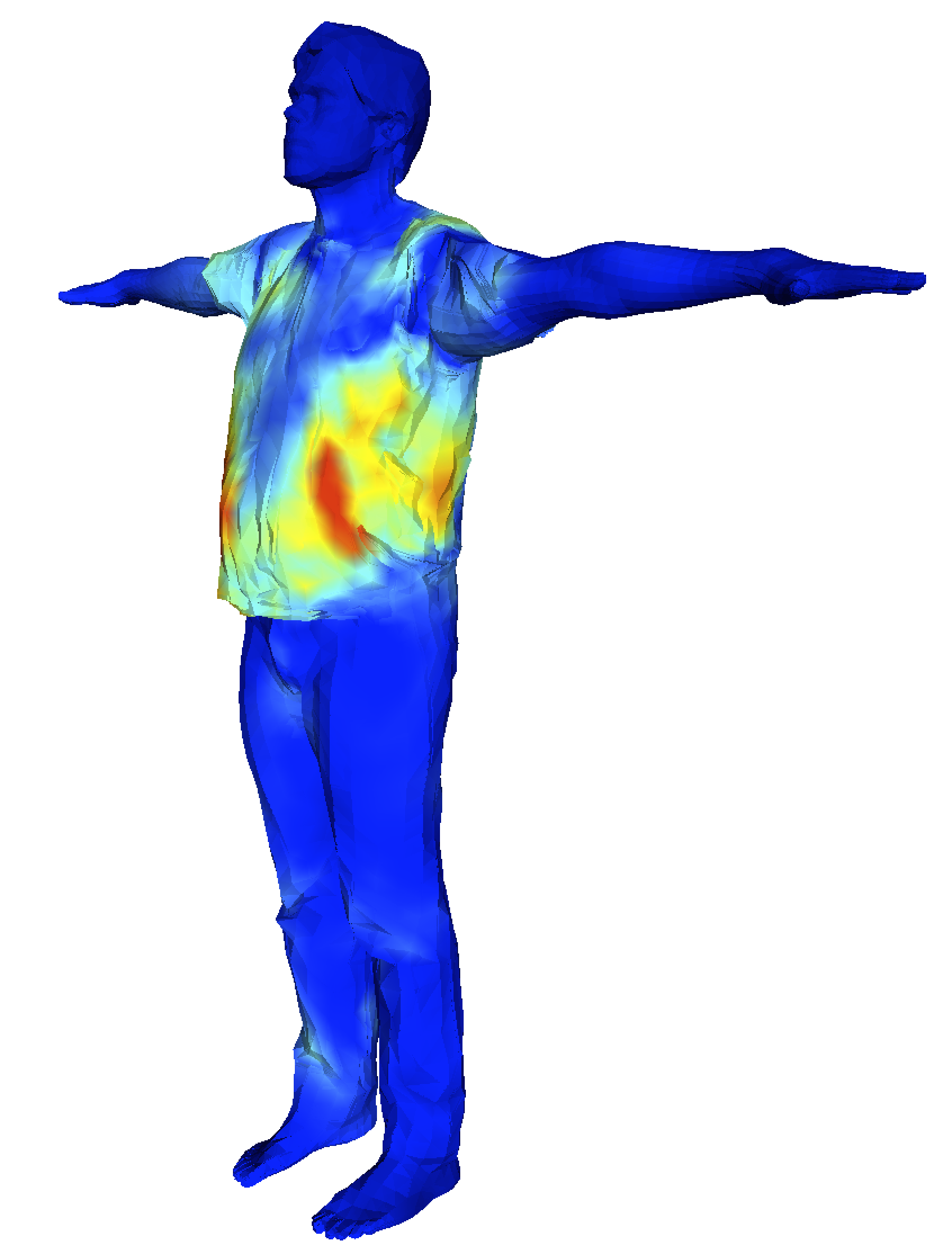}\llap{c~} &
\includegraphics[height=2.8cm]{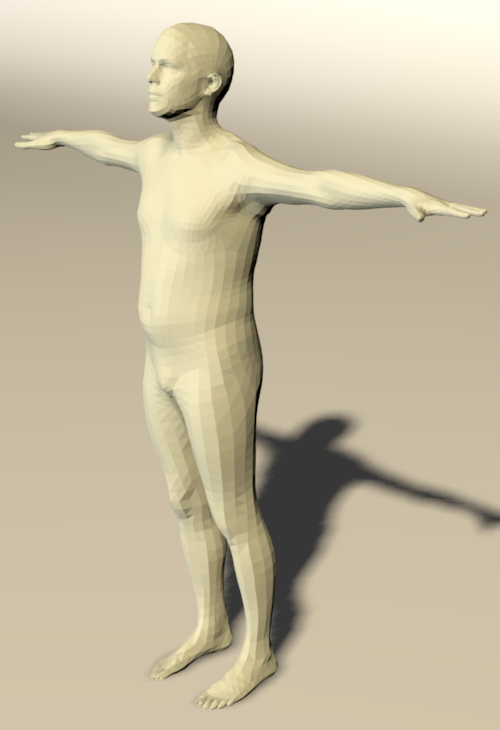}\llap{d~} &
\includegraphics[height=2.8cm]{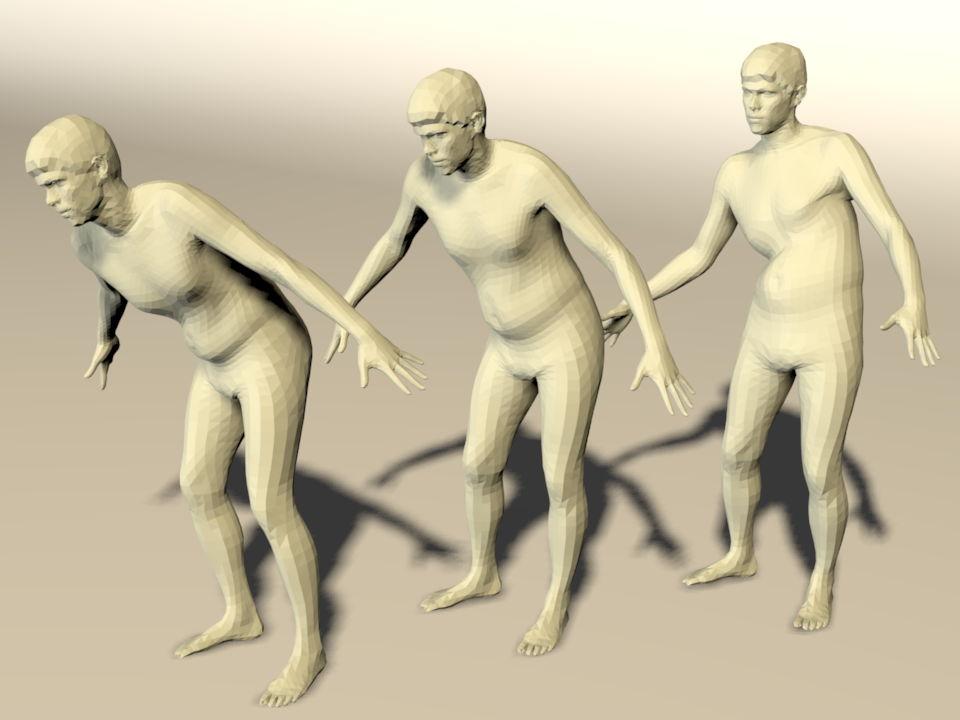}\llap{e~} \\
\end{tabular}
\caption{a) Cloth alignments b) Unposed alignments c) Fusion scan d) Fusion shape e) Posed and tracked shape. Overview: three example frames are shown. Notice the match in the cloth wrinkles between posed a) and unposed b) alignments.
Different time frames provide different constraints in the unposed space. The fusion scan is the union of the frame wise unposed alignments. Color code indicates variance for that region. From the fusion scan c) we obtain the fusion shape d).}
\label{fig:outline}
\end{figure*}

SMPL \cite{loper2015smpl} is a body model that uses a learned rigged
template $\mat{T}$ with $N=6890$ vertices.
The vertex positions of SMPL are adapted according to identity-dependent shape parameters and the skeleton pose.
The skeletal structure of the human body is modeled with a kinematic chain consisting of rigid bone segments linked by $n=24$ joints. 
Each joint is modeled as a ball joint with 3 rotational Degrees of Freedom (DoF), parametrized with exponential coordinates $\vect{\omega}$.
Including translation, the pose $\pose$ is determined by a pose vector
of $3\times{24} + 3 = 75$ parameters.


To model shape and pose dependent deformations SMPL
modifies the template in an additive way and predicts the joint 
locations from the deformed template. 
The model, $\smpl(\shape,\pose)$ is then
\begin{eqnarray}
\smpl(\shape,\pose) & = & W
(T(\shape,\pose),J(\shape),\pose,\bweights ) \\
T(\shape, \pose)
& = & \vt+\shapebs+\posebs
\label{eq:SMPL}
\end{eqnarray}
where $W(\vt, \pose, \joints) :
\R^{3N}\times{\R}^{|\theta|}\times{\R}^{3K}\mapsto \R^{3N}$ is a
linear blend skinning function that  takes
vertices in the rest pose $\vt$, joint locations $\joints$, a pose $\pose$, and the
blend weights $\bweights$, and returns the posed vertices.
The parameters $\shapebs$ and $\posebs$ are vectors of 
vertex offsets from the template.
We refer to these as shape and pose blend shapes respectively.
We use $\mathcal{M}$ to refer to the mesh produced by SMPL.
Note that this is different from $\smpl$, which only refers to the vertices.
See \cite{loper2015smpl} for more details.


\section{Method}
\label{sec:method}
\setlength{\parskip}{-0.05em}
\setlength{\belowdisplayskip}{3pt} 
\setlength{\abovedisplayskip}{3pt}

\begin{figure}
\centering
\includegraphics[width=0.23\columnwidth]{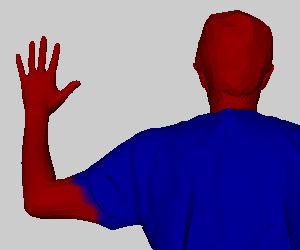}\llap{\color{white}a~}
\includegraphics[width=0.23\columnwidth]{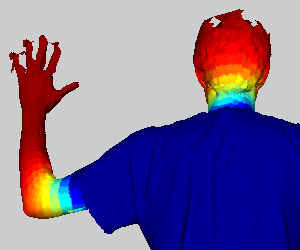}\llap{\color{white}b~} 
\includegraphics[width=0.23\columnwidth]{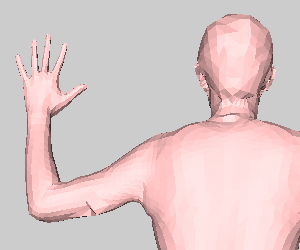}\llap{c~} 
\includegraphics[width=0.23\columnwidth]{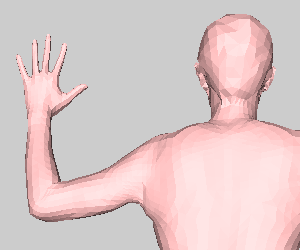}\llap{d~} \\
\caption{Skin term weights.
a) alignment segmentation (red: skin, blue: cloth) 
b) geodesic distance to the closest cloth vertex on the alignment
c) broken result with unsmooth neck and arms 
d) smooth result.}
\label{fig:geodesic}
\end{figure}


Our goal is to estimate the naked shape and pose of a subject from a sequence of clothed scans $\{ \scan \}_k$.
If the scans have color information, we use it to split
the scan vertices into two sets:
the skin ($\scan_\mathrm{skin}$) and the cloth ($\scan_\mathrm{cloth}$),
otherwise we consider all vertices as cloth ($\scan_\mathrm{cloth} = \scan$).
Here we use the segmentation method in~\cite{Pons-Moll:Siggraph2017}, 
see Sup.~Mat. for more details.
The outputs of our method are:
a personalized static template shape $\vtFusion$,
the per frame poses $\theta_k$, and the per frame detailed template shapes $\vtEst^k$. 
Ideally, pose dependent shape changes should be explained by $\vtFusion$ and the pose deformation model; however, in practice models deviate from real data.
Therefore, we allow our result $\vtEst^k$ to slightly vary over time.
This allows us to capture time changing details, e.g. facial
details, present in the data, which the model can not represent.

Given a single scan we obtain the shape by minimizing a $\emph{single-frame objective}$ function (\sectref{sec:single_frame})
that constrains the scan cloth points to be outside of the body, and penalizes deviations from the body
to skin parts. 
However, estimating the shape from a single scan is an under-constrained problem. 
Fortunately, when all the information in a sequence is considered,
 the underlying shape is more constrained,
as different poses will make the cloth tight to the body in different parts. 
In order to exploit such rich temporal information
we first bring all input scans into correspondence.
As a result we obtain a set of posed registrations
and unposed template registrations (see \figref{fig:outline} a and b).
The union of the unposed templates creates the \emph{fusion scan} (\figref{fig:outline} c).
We use it to estimate a single shape, that we call the \emph{fusion shape}
(\figref{fig:outline} d).
Since all temporal information is fused into a single fusion scan,
we can estimate the fusion shape using the same single-frame objective function.
Using the fusion shape template as a prior, we can accurately estimate the pose and shape
of the sequence.
In~\figref{fig:outline} we show the different steps of the method. 
The results of each stage are obtained using variants of the same single-frame objective. 

\subsection{Single-Frame Objective}
\label{sec:single_frame}
We define the single-frame objective function as:
\begin{align*}
E(\vtEst, \smpl(\shape, 0), \pose; \scan) &=  & \lambda_\mathrm{skin} E_\mathrm{skin} + 
E_\mathrm{cloth} \\
 && + \lambda_\mathrm{cpl} E_\mathrm{cpl} + \lambda_\mathrm{prior} E_\mathrm{prior},
\end{align*}
where $E_\mathrm{skin}$ is the skin term, $E_\mathrm{cloth}$ is the cloth term,
$E_\mathrm{cpl}$ is the model coupling term and $E_\mathrm{prior}$ includes prior terms for pose, shape, and translation.
$\smpl(\shape, 0) = \vt + \shapebs$; $\vt$ is the default template of the SMPL model, and $\shape$ are the 
coefficients of the shape space, see~\eqnref{eq:SMPL}.
Next we describe each of the terms.

\paragraph{Skin term:} 
We penalize deviations from the model to scan points labeled as skin $\vect{s}_i \in \mathcal{S}_\mathrm{skin}$ (see \figref{fig:geodesic}).
A straightforward penalty applied to only skin points creates
 a discontinuity at the boundaries, which leads to poor results~(\figref{fig:geodesic} c).
In order to make the cost function smooth, we first compute the geodesic distance of a point in the alignment to the closest cloth point, and we
apply a logistic function to map geodesic distance values between 0 and 1 (\figref{fig:geodesic} b).
We name this function $g(\vect{x}):\R^3 \mapsto [0,1]$.
The resulting value is propagated to the scan points by nearest distance,
and used to weight each scan residual.
This way, points close to skin-cloth boundaries have a smooth decreasing weight.
This effectively makes the function smooth and robust to inaccurate segmentations (\figref{fig:geodesic} d). 
\begin{equation}
E_\mathrm{skin}(\vtEst,\pose; \scan) = \sum_{\vect{s_i} \in \skin} g(\vect{s}_i) \rho(\mathrm{dist}(\scanvertex_i, \smplsurf(\vtEst,\pose))),
\end{equation}
where $\mathrm{dist}$ is a point to surface distance, and $\rho(\cdot)$ is Geman-McClure penalty function. Note that
$\mathrm{dist}()$ is computing the closest primitive on the mesh $\smplsurf(\vtEst,\pose)$, triangle, edge or point;
analytic derivatives are computed accordingly in each case.
\paragraph{Cloth term:} 
The cloth objective consists of two terms: $E_\mathrm{cloth} = \lambda_{\mathrm{outside}}E_\mathrm{outside} + \lambda_{\mathrm{fit}}E_\mathrm{fit}$. The \emph{outside} term penalizes cloth points penetrating the mesh and the \emph{fit}
term encourages the mesh to remain close to the cloth surface. This is in contrast to previous work \cite{yang2016estimation}
that assumes a closed scan and pushes the model inside. Since scans are not closed surfaces we just penalize cloth points penetrating our 
closed registration surface. Therefore, the approach is general to point clouds.  
The outside term is mathematically the sum of penalties for every scan point labeled as cloth $\scanvertex \in \cloth$
that penetrates the shape mesh: 
\begin{equation}
E_\mathrm{outside}(\vtEst,\pose; \scan) = \sum_{\scanvertex_i \in{\cloth}} \delta_{i} \mathrm{dist}(\scanvertex_i, \smplsurf(\vtEst,\pose))^2,
\end{equation}
where $\delta_{i}$ is an indicator function returning 1 if the scan point $\scanvertex_{i}$ lies inside the mesh and 0 otherwise. 
The activation $\delta_{i}$ is easily obtained by computing the angle between the mesh surface normal and
the vector connecting the scan vertex and the closest point in the mesh.

\begin{figure}
\centering
\includegraphics[width=0.49\columnwidth]{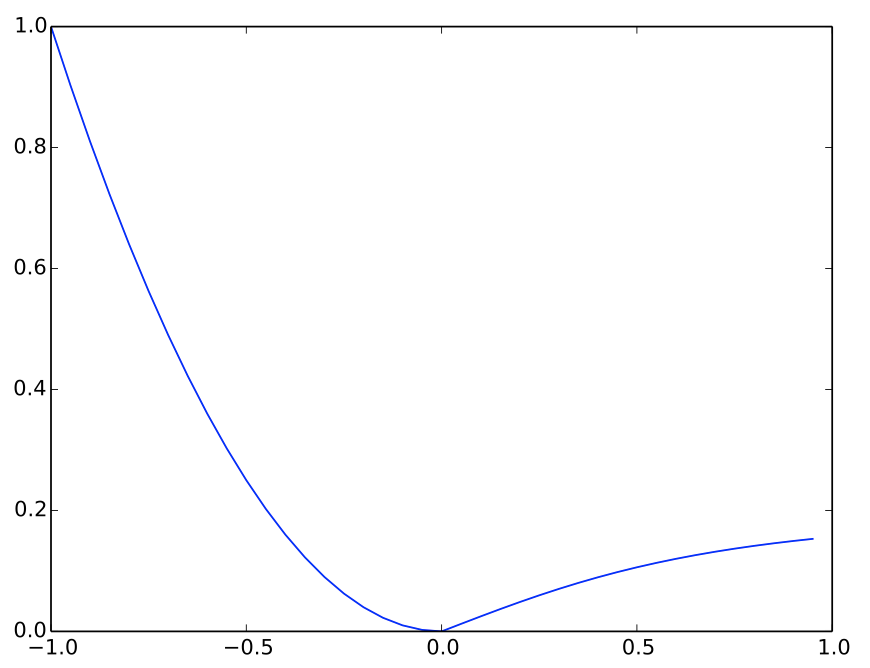}
\includegraphics[width=0.49\columnwidth]{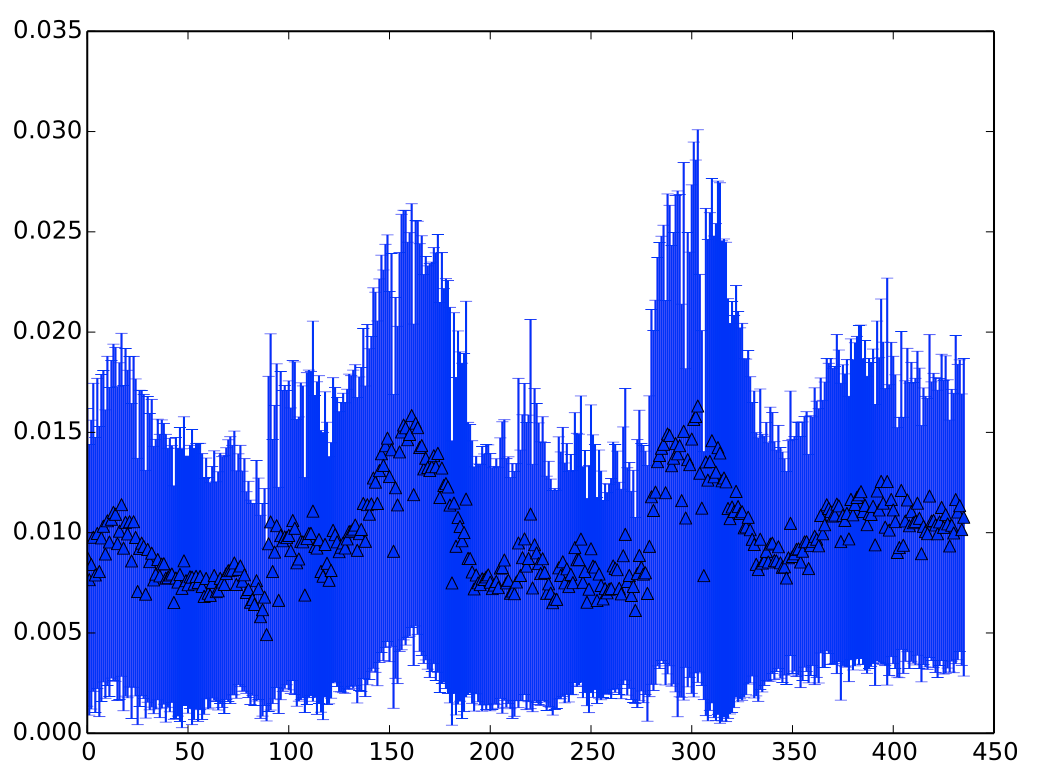}
\caption{Left: Cloth term. The x-axis is the signed distance between $\scanvertex \in \cloth$ and $\smplsurf(\vtEst,\pose)$. Points inside (negative) have a quadratic penalty, while points outside are penalized using a robust Geman-McClure function.
Right: Root mean squared error and standard deviation between single-frame estimations and the ground truth.
Results have significant dispersion depending on pose.
(Results for subject 00005, motion ``hips'' and clothing style ``soccer''.)}
\label{fig:cloth_objective}
\end{figure}

Minimization of the outside term alone can make the shape excessively thin.
Hence, the fit term $E_\mathrm{fit}$ is used to maintain the volume of the naked model.
Every cloth scan vertex pays a penalty if it deviates from the body. 
Since we want to be robust to wide clothing, 
we define $E_\mathrm{fit}$ as a Geman-McClure cost function.
With this robust cost function, 
points far away (\eg points in skirt or wide jacket)
 pay a small nearly-constant penalty. 
The resulting cloth term is illustrated in the left part of \figref{fig:cloth_objective}.

\paragraph{Coupling term:} 
Optimizing only $E_\mathrm{skin}$ and $E_\mathrm{cloth}$ results in very unstable
 results because no human anthropometric constraints are enforced.
Therefore, we constrain the template $\vtEst$ to remain close to the statistical shape body model
\begin{equation}
E_\mathrm{cpl}(\vtEst, \smpl(0,\shape)) =  \| \mathrm{diag}(\vect{w}) ( \mathbf{T}_\mathrm{Est,e} - \smpl(0, \shape)_e)\|^2
\end{equation}
where the diagonal matrix $\mathrm{diag}(\vect{w})$ simply 
increases the coupling strength for parts like hands and feet where the scans are noisier. Coupling is performed
on edges indicated by underscript $e$.
Since we are jointly optimizing $\vtEst$, and $\shape$, the model of the shape
is pulled towards $\vtEst$ and vice versa. The result of the optimization is a detailed estimation $\vtEst$
and a model representation of the shape $\shape$. 
\paragraph{Prior term:} 
The pose is regularized with a Gaussian prior computed from the pose training set of \cite{loper2015smpl}. 
Specifically we enforce a Mahalanobis distance prior on the pose: 
\begin{equation}
E_\mathrm{prior}(\theta) = (\pose-\mu_{\pose})^T \Sigma^{-1}_{\pose}(\pose-\mu_{\pose})
\end{equation}
where the mean $\mu_{\pose}$ and covariance $\Sigma_{\pose}$ are computed from the pose
training set. A similar prior can be enforced on the shape space
coefficients $\shape$ but we found it did not make
a significant difference.

To optimize the single-frame objective we compute the derivatives using the 
autodifferentiation tool Chumpy \cite{chumpy}.
We use the ``dogleg'' gradient-based descent minimization method \cite{Nocedal2006NO}.

\subsection{Fusion Shape Estimation}
\label{sec:fusion_shape}
The problem with the single-frame objective is two fold: the temporal information is disregarded and
the frame wise shape changes over time depending on the pose. This can be seen in the right part of \figref{fig:cloth_objective}.
The straightforward approach is to extend the single-frame objective 
to multiple frames and optimize jointly
a single $\vtEst$, $\shape$ and the $\nframes$ poses $\{ \pose^k\}_{k=1}^{\nframes}$. 
Unfortunately, our scans have around $150,000$ points,
and optimizing all poses jointly makes the optimization
highly inefficient and memory intensive. Furthermore, slight miss-alignments in pose make the shape shrink too much. 
Hence, we propose an effective and more efficient solution. We first sequentially register all the scans to a single
clothed template. For registration we use the single-frame objective function with no cloth term.
From this we obtain a template clothed per frame $\vtCloth^k$. The
interesting thing is that the set of $\vtCloth^k$ templates contain
the non-rigid cloth motion with the motion due to pose factored out, see \figref{fig:outline}. The naked shape should lie inside 
all the clothed templates. Hence we gather all templates and treat them as a single point cloud
that we call the fusion scan $\fusionscan=\{ \vtCloth^k \}_{k=1}^{\nframes}$. 
Hence, we can easily obtain a single shape estimate by using again the single-frame objective
\begin{equation}
\vtFusion = \argmin{\vtEst, \shape}\quad E(\vtEst, \smpl(\shape, 0), 0; \fusionscan) .
\end{equation}
The obtained fusion shapes are already quite accurate because the fusion scan carves the volume where
the naked shape should lie. 
\subsection{Pose and Shape Tracking}
\label{sec:tracking}
Finally we use the fusion shape to perform tracking regularizing the estimated shapes to remain close
to the fusion shape. 
We achieve that by coupling the estimations towards the fusion shape instead of
the SMPL model shape space. So the coupling term is now 
$E_\mathrm{cpl}(\vtEst, \smpl(0,\shape)) \mapsto E_\mathrm{cpl}(\vtEst, \vtFusion)$. Detailed shapes are obtained minimizing
\begin{equation}
\vtEst^k = \argmin{\vtEst, \pose}\quad E(\vtEst, \vtFusion,\pose .
\scan^k) .
\end{equation}

\section{Datasets}
\label{sec:dataset}
In this section we present our new \dataset dataset.
We start by introducing the previous dataset. 

\subsection{Existing Dataset}
The INRIA dataset \cite{yang2016estimation} consists
of sequences of meshes obtained by applying a visual hull reconstruction to a 68-color-camera (4M pixels) system at 30fps.
The dataset includes sparse motion capture (MoCap) data of 6 subjects (3 female, 3 male) captured in 3 different motions and 3 clothing styles each.
The texture information of the scans is not available.
\figref{fig:Inria_dataset} a) and b) show frames from the dataset.
The ``ground truth shape" of a subject is estimated by fitting the
S-SCAPE \cite{jain2010} model to the ``tight clothes'' sequence.

As shown in \figref{fig:Inria_dataset} c) and d),
their statistical body model does not capture the individual details of the human shape.
The main drawback of this ``ground truth shape'' is that it biases the
evaluation to the model space.
All recovered details, that fall outside the model, will be penalized in the quantitative evaluation.
Alternatively, one could compare the obtained shape directly with the visual hull. 
Unfortunately, visual hulls are not very accurate, sometimes
over-estimating, sometimes under-estimating the true shape.
While relevant for qualitative evaluation of the shape estimates,
we believe that this dataset is limited for quantitative evaluation.
%
%
%
This motivated us to create BUFF, which preserves details and 
allows quantitative evaluation of the shape estimation.

\newcommand{\resinriadatasetwidth}{0.23\columnwidth}
\begin{figure}
\centering
\includegraphics[width=\resinriadatasetwidth, trim={100 00 260 140}, clip]{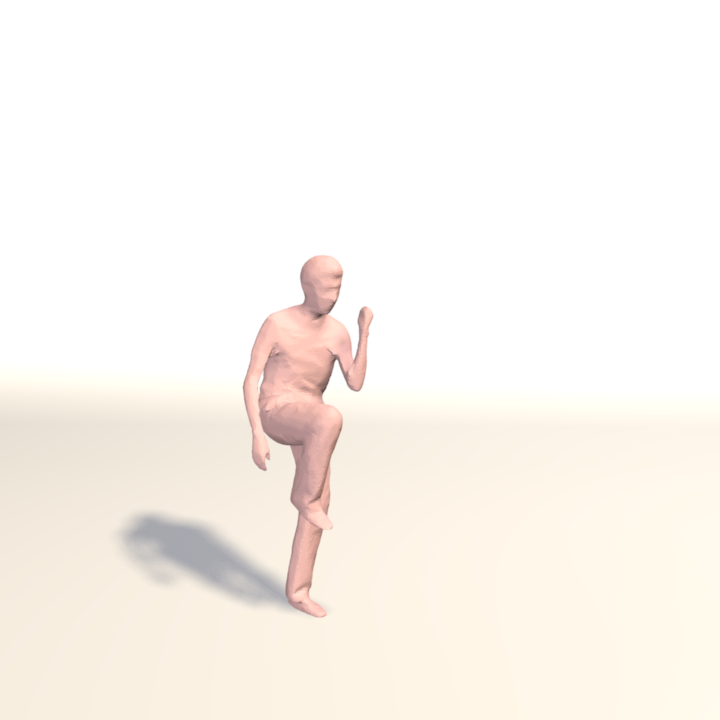}\llap{a)}
\includegraphics[width=\resinriadatasetwidth]{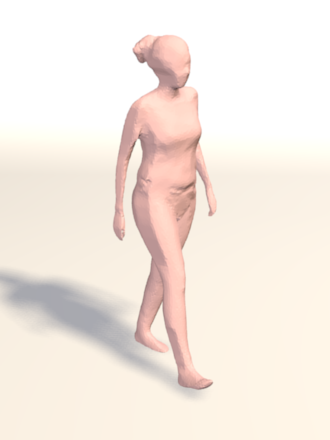}\llap{b)}
\includegraphics[width=\resinriadatasetwidth]{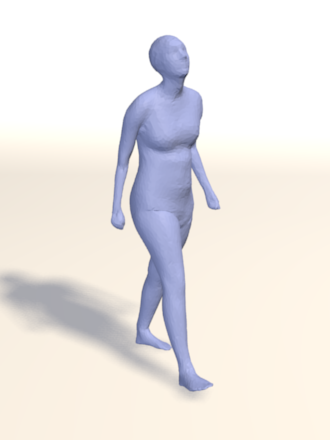}\llap{c)}
\includegraphics[width=\resinriadatasetwidth]{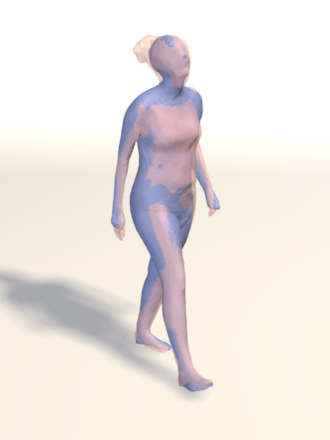}\llap{d)}
\caption{INRIA Dataset: a) and b) scan samples; c) estimated ``ground truth" shape for b);
d) overlay of b) and c).}
\label{fig:Inria_dataset}
\end{figure}


\subsection{\dataset}
To create BUFF, we  use  a  custom-built  multi-camera  active  stereo
system  (3dMD  LLC,  Atlanta,  GA)  to  capture  temporal sequences
of full-body 3D scans at 60 frames per second.
The system uses 22 pairs of stereo cameras, 22 color cameras, 34
speckle projectors and arrays of white-light LED panels.
The projectors and LEDs flash at 120fps to alternate between stereo 
capture and color capture.
The projected texture pattern makes stereo matching more accurate,
dense,  and  reliable  compared  with  passive  stereo  methods.
The stereo pairs are arranged to give full body capture for a range of
activities, enabling us to capture people in motion.
The system outputs 3D meshes with approximately 150K vertices on
average.

\begin{figure}
\centering
\includegraphics[width=0.99\columnwidth]{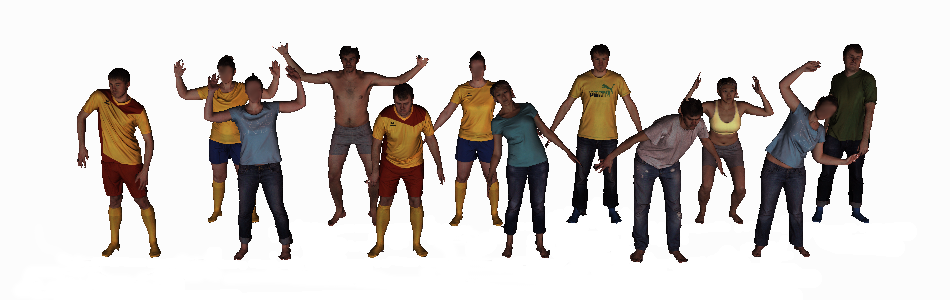}
\caption{BUFF Dataset: To validate our method we captured a new
  dataset including 6 subjects wearing different clothing styles and
  different motion patterns.
}
\label{fig:NakeCap_dataset} 
\end{figure}

\dataset consists of 6 subjects, 3 male and 3 female wearing 2 clothing styles: a) t-shirt and long pants and b) a soccer outfit, see \figref{fig:NakeCap_dataset}.
The sequence lengths range between 4 to 9 seconds (200-500 frames) totaling 13,632 3D scans.
%

As shown by previous state of the art methods \cite{bualan2008naked}, skin color is a rich source of information.
We thus include texture data in our dataset.
%
All subjects gave informed written consent before participating in the
study. One subject did not give permission to release their data for
research purposes.
Consequently, the public BUFF dataset consists of 11,054 scans.

\subsubsection{Computing the Ground Truth Shapes}
\label{sec:gt_naked}

\begin{figure}
\centering
\includegraphics[width=0.24\columnwidth]{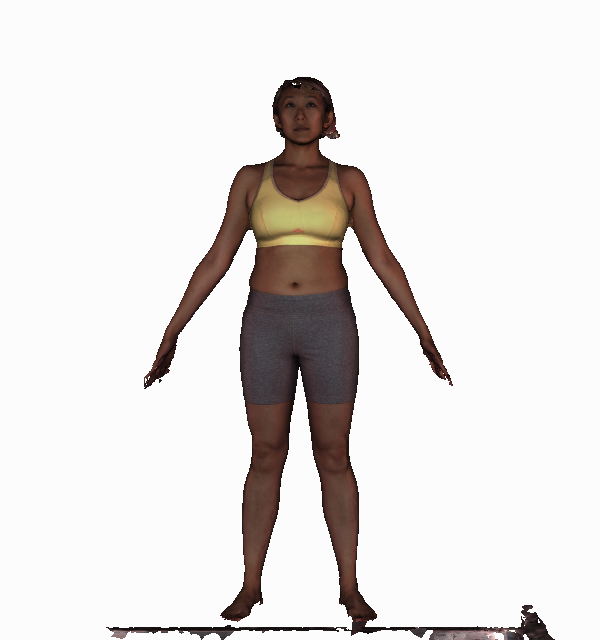}
\includegraphics[width=0.24\columnwidth]{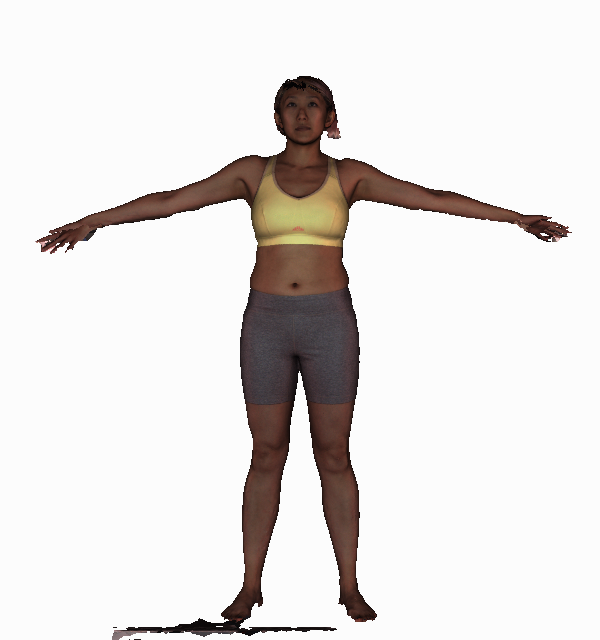}
\includegraphics[width=0.24\columnwidth]{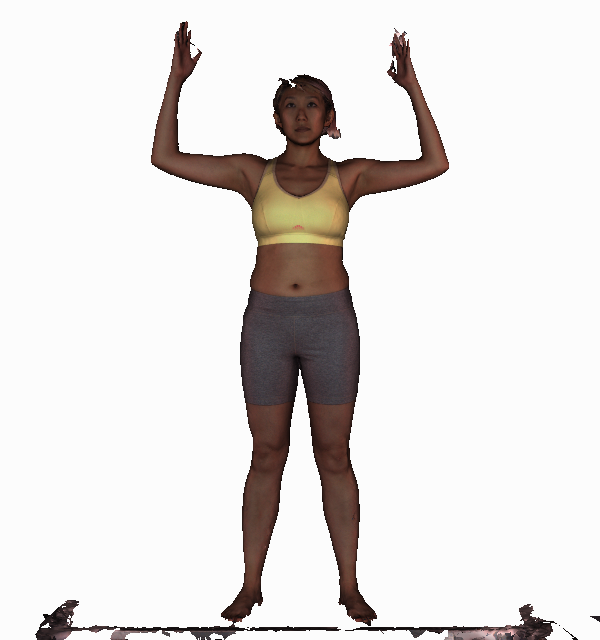}
\includegraphics[width=0.24\columnwidth]{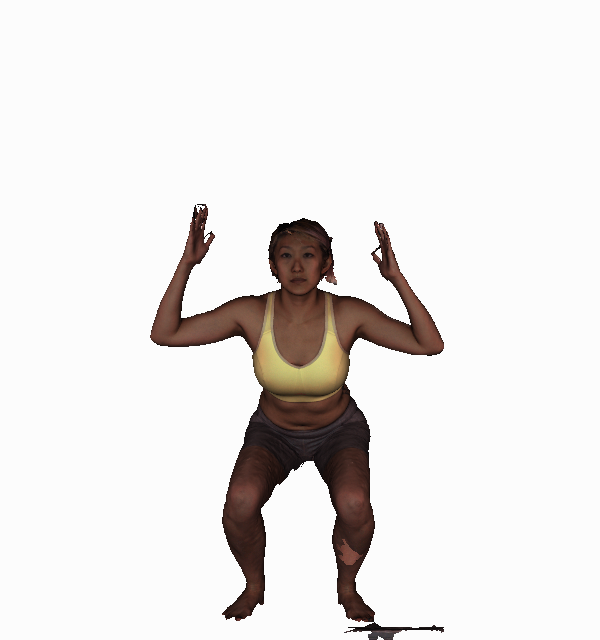} \\
\includegraphics[width=0.25\columnwidth]{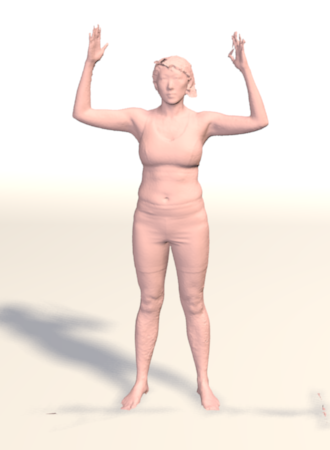}
\includegraphics[width=0.25\columnwidth]{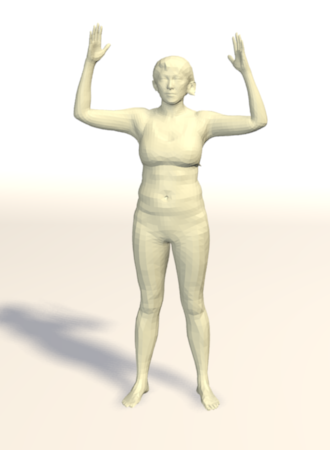}
\includegraphics[width=0.25\columnwidth]{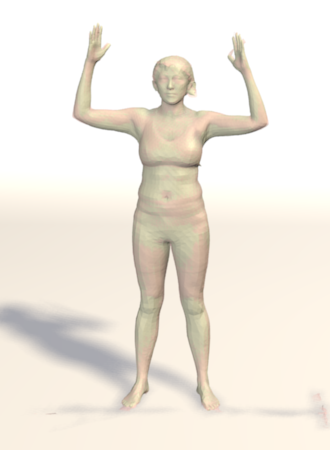} \\
\caption{Top row: Subject 03223 performing the ``A-T-U-Squat'' motion in ``minimal clothing''. 
These scans are used to compute the ground truth MCS  $\vtGT$.
Bottom row: triplet of scan, estimated ground truth model and both overlayed (frame 000150).
The proposed ground truth shape captures the details present in the scan point cloud.}
\label{fig:minimal_ATU}
\end{figure}

In order to estimate the ``ground truth'' shapes in our dataset we capture
 the subjects in ``minimal clothing'' (tight fitting sports underwear).
Participants performed an ``A-T-U-Squat'' motion (first row of \figref{fig:minimal_ATU}).
For all frames, we use our method to fit the data considering all vertices as ``skin'' (see \sectref{sec:single_frame}).
We obtain $N$ template meshes $\vt^i$, which do not perfectly match, because 
the pose and the shape are not perfectly factorized in the SMPL model \cite{loper2015smpl}.
We define the $\vtGT$ as the mean of the estimates of all frames.

We quantitatively estimated the accuracy of our ``ground truth'' MCS estimations.
More than half of the scan points are within 1.5mm distance of $\vtGT$ and $80\%$ closer than 3mm.
Because the scan point cloud has some noise (e.g. points of the scanning platform, poorly reconstructed hands, hair,...),
we believe the computed $\vtGT$ provides an accurate explanation of
the subjects ``minimally clothed shape''.
In the bottom row of \figref{fig:minimal_ATU} we qualitatively show the visual accuracy of the computed ground truth MCS.

\section{Experiments}
\label{sec:experiments}
\newcommand{\resthreewidth}{0.3\columnwidth}

\begin{figure}
\centering
\includegraphics[width=\resthreewidth]{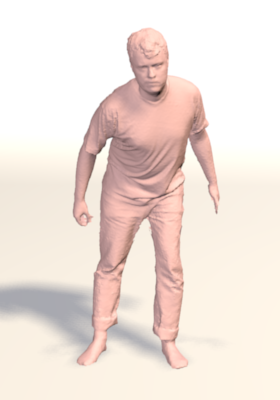} 
\includegraphics[width=\resthreewidth]{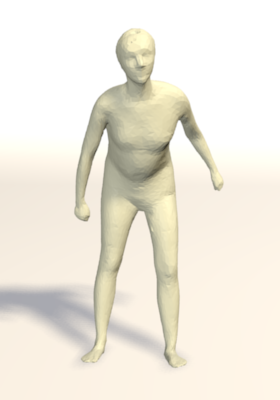} 
\includegraphics[width=\resthreewidth]{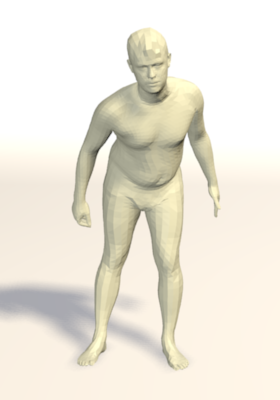} \\
\vspace{-3pt}
\includegraphics[width=\resthreewidth]{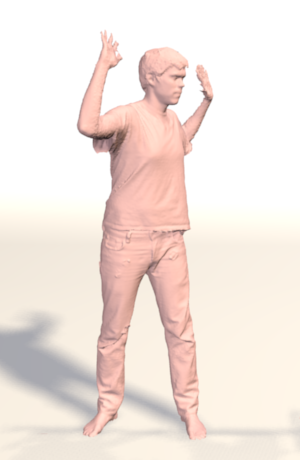} 
\includegraphics[width=\resthreewidth]{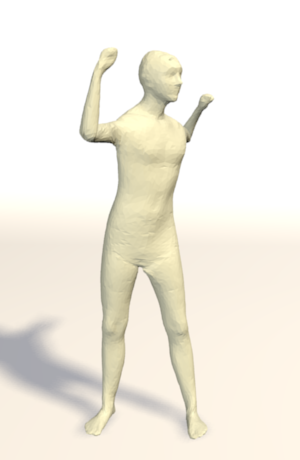} 
\includegraphics[width=\resthreewidth]{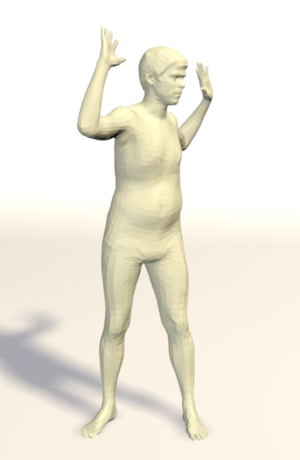}\\
\vspace{-3pt}
\includegraphics[width=\resthreewidth]{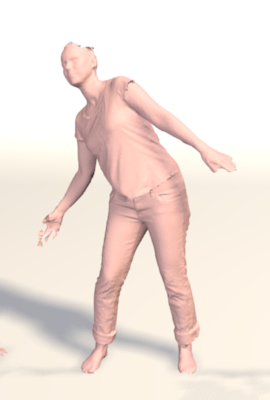} 
\includegraphics[width=\resthreewidth]{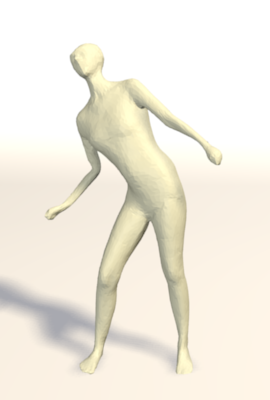} 
\includegraphics[width=\resthreewidth]{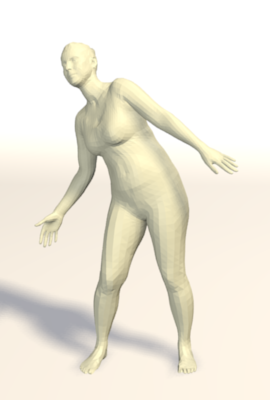} \\
\vspace{-3pt}
\caption{Qualitative pose estimation results on BUFF dataset. Left to right: scan, Yang \etal~\cite{yang2016estimation}, our result.}
\vspace{-10pt}
\label{fig:NakeCap_results_pose} 
\end{figure}

\newcommand{\reswidth}{0.24\columnwidth}
\begin{figure}
\centering
\includegraphics[width=\reswidth]{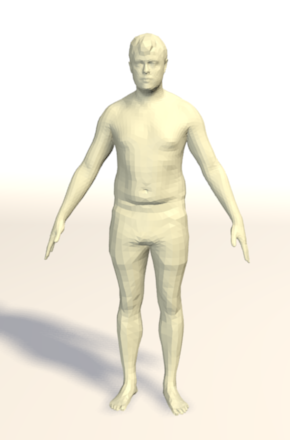}
\includegraphics[width=\reswidth]{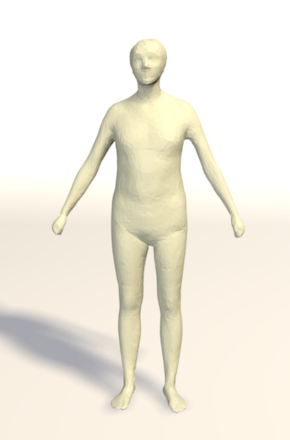}
\includegraphics[width=\reswidth]{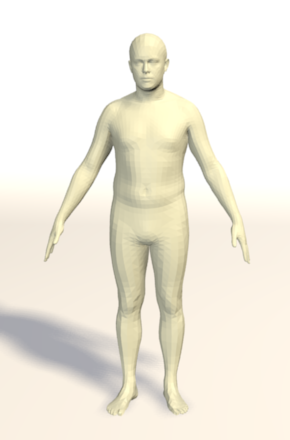}
\includegraphics[width=\reswidth]{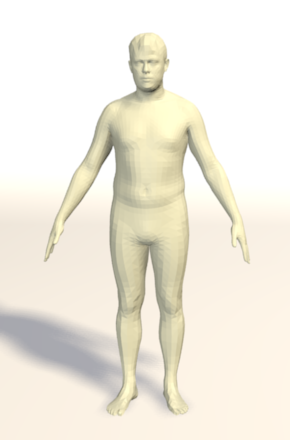}\\
\vspace{-5pt}
\includegraphics[width=\reswidth]{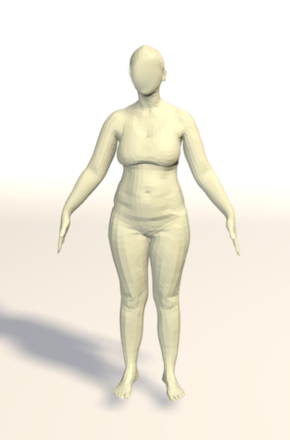}
\includegraphics[width=\reswidth]{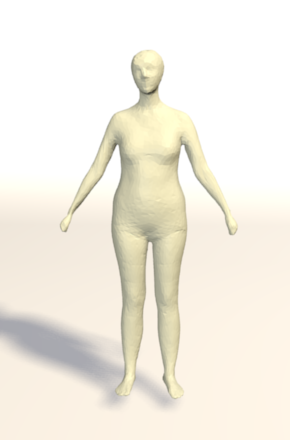}
\includegraphics[width=\reswidth]{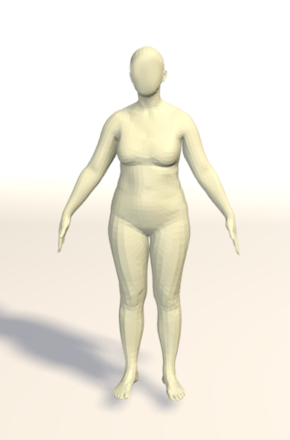}
\includegraphics[width=\reswidth]{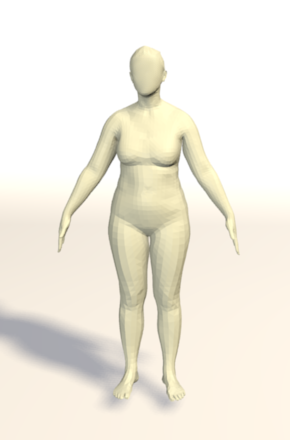}\\
\vspace{-5pt}
\includegraphics[width=\reswidth]{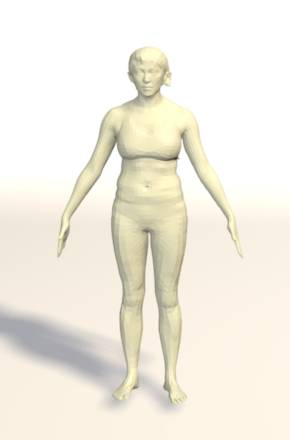}
\includegraphics[width=\reswidth]{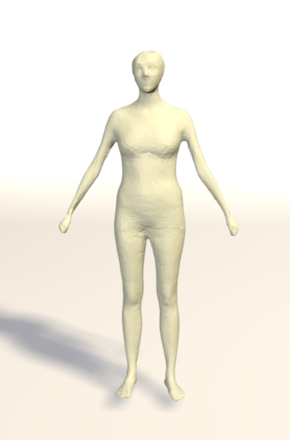}
\includegraphics[width=\reswidth]{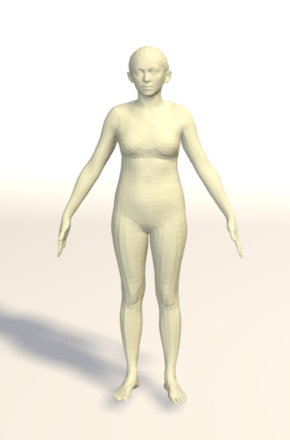}
\includegraphics[width=\reswidth]{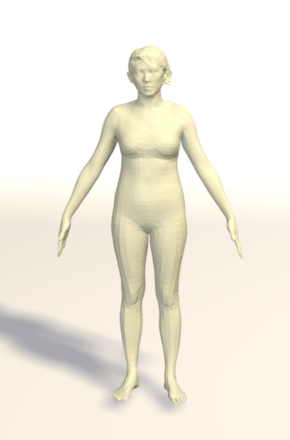}
\vspace{-5pt}
\caption{Qualitative shape estimation results on BUFF dataset. Left to right: ground truth shape, Yang \etal~\cite{yang2016estimation},
fusion shape (ours), detailed shape (ours).
}
\label{fig:NakeCap_results_shape} 
\end{figure}

\begin{figure}
\centering
\includegraphics[width=0.49\columnwidth]{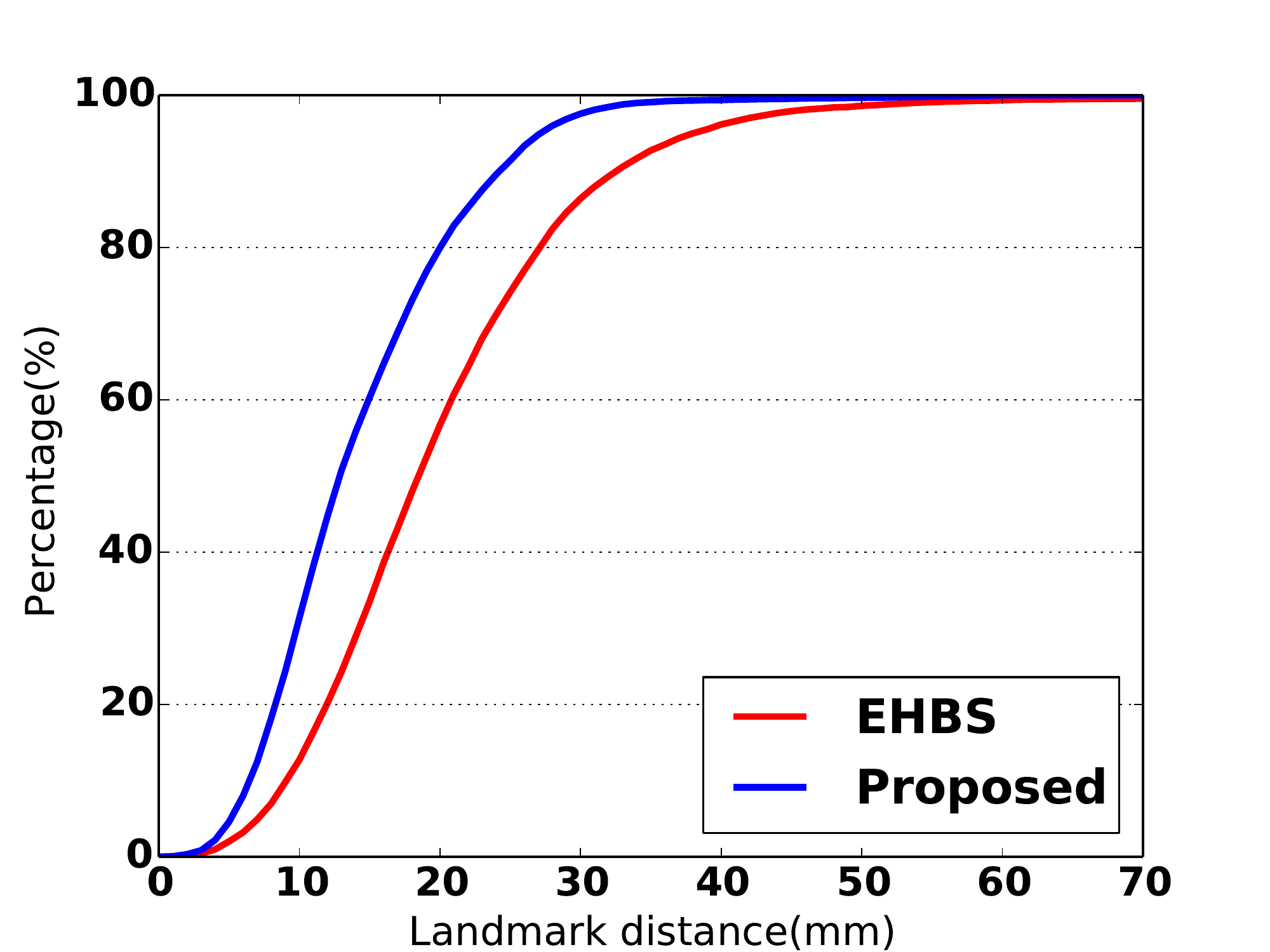}
\includegraphics[width=0.49\columnwidth]{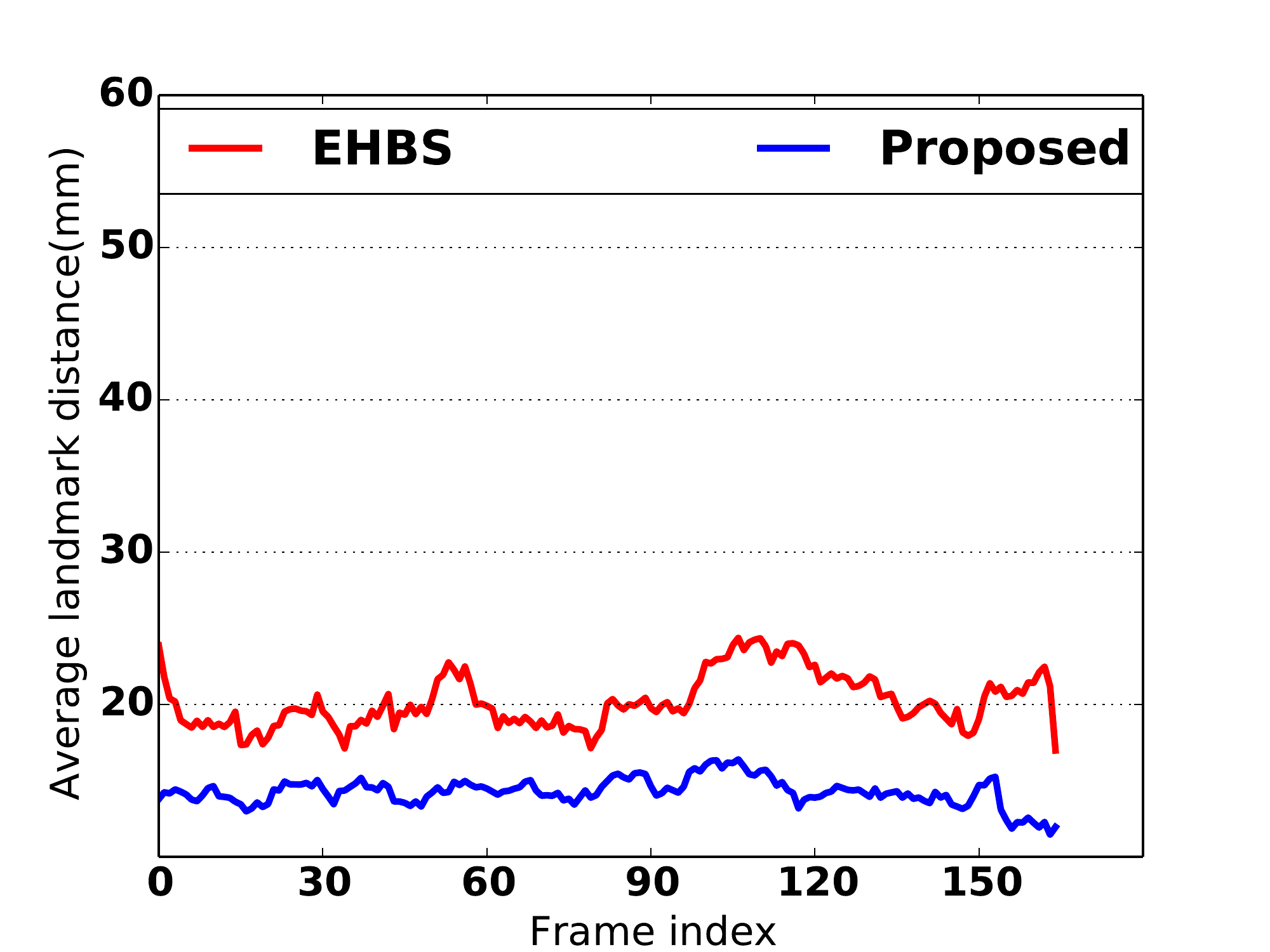}
\caption{Pose estimation accuracy on INRIA dataset.
Left: Percentage of landmarks with error less than a given distance (horizontal axis) in mm.  Right: per frame average landmark error. EHBS is \cite{yang2016estimation}.
}
\label{fig:Inria_results_pose} 
\end{figure}
In this section we present the evaluation measures and the obtained qualitative and quantitative results.
\begin{table*}
\centering
{\small
\begin{tabular}{|l|cccccc||ccccc|c|} 
\hline 
& \multicolumn{6}{|c||}{t-shirt, long pants}
& \multicolumn{5}{c|}{soccer outfit}
& \multicolumn{1}{c|}{Avrg.}   \\ 
\hline 
 {\em tilt twist left} 
  & 00005 & 00096 & 00032 & 00057 & 03223 & 00114 
  & 00005 & 00032 & 00057 & 03223 & 00114 & Avrg.\\ 
\hline 
Yang \etal~\cite{yang2016estimation}  
& 17.29 & 18.68 & 13.76 & 17.94 & 17.90 & 15.42 
& 16.77 & 16.96 & 18.52 & 20.41 & 16.40 & 17.27\\ 
fusion mesh 
& 2.58 & 2.89 & 2.39 & 2.53 & 2.43 & 2.38 
& 2.50 & 2.63 & 2.37 & 2.28 & 2.28 & 2.47\\ 
detailed 
& \bf{2.52} & \bf{2.83} & \bf{2.36} & \bf{2.44} & \bf{2.27} & \bf{2.31} 
& \bf{2.44} & \bf{2.59} & \bf{2.28} & \bf{2.17} & \bf{2.23} & \bf{2.40}\\ 
\hline 
 \hline 
 {\em hips} 
  & 00005 & 00096 & 00032 & 00057 & 03223 & 00114 
  & 00005 & 00032 & 00057 & 03223 & 00114 & Avrg.\\ 
\hline 
Yang \etal~\cite{yang2016estimation}
  & 21.02 & 21.66 & 15.77 & 17.87 & 21.84 & 18.05 
  & 22.52 & 16.81 & 19.55 & 22.03 & 17.54 & 19.51\\ 
fusion mesh & 2.81 & 2.71 & 2.66 & 2.66 & 2.54 & 2.65 
          & 2.65 & 2.63 & 2.58 & 2.50 & 2.57 & 2.63\\ 
detailed & \bf{2.75} & \bf{2.64} & \bf{2.63} & \bf{2.55} & \bf{2.40} & \bf{2.56} 
         & \bf{2.58} & \bf{2.59} & \bf{2.50} & \bf{2.38} & \bf{2.51} & \bf{2.55}\\ 
\hline 
 \hline 
 {\em shoulders mill} 
  & 00005 & 00096 & 00032 & 00057 & 03223 & 00114 
  & 00005 & 00032 & 00057 & 03223 & 00114 & Avrg.\\ 
\hline 
Yang \etal~\cite{yang2016estimation}  
& 18.77 & 19.02 & 18.02 & 16.50 & 18.15 & 14.78 
      & 18.74 & 17.88 & 15.80 & 19.47 & 16.37 & 17.59\\ 
fusion mesh 
& 2.56 & 2.92 & 2.74 & 2.46 & 2.42 & 2.69 
& 2.89 & 2.87 & 2.37 & 2.44 & 2.58 & 2.63\\ 
detailed 
& \bf{2.49} & \bf{2.85} & \bf{2.72} & \bf{2.37} & \bf{2.26} & \bf{2.59} 
& \bf{2.83} & \bf{2.82} & \bf{2.28} & \bf{2.33} & \bf{2.51} & \bf{2.55}\\ 
\hline 
\end{tabular}
\vspace{2mm} 
} 
\caption{Numerical results for the estimated naked shapes. 
We report the root mean squared error in millimeters of point to surface distance 
between the posed GT mesh and the method result. 
The best value is highlighted in bold.
} 
\label{tab:quantitative_results} 
\end{table*}

\subsection{Evaluation on Previous Datasets}
\label{sec:evaluation_inria_dataset}
We evaluate our results quantitatively on pose estimation,
and qualitatively on shape estimation in the INRIA dataset \cite{yang2016estimation}.
We estimated the shape for all tight clothes sequences. To initialize the pose we use the automatically computed landmarks of \cite{zuffi2015stitched}.
We compare the MoCap marker locations to the corresponding vertices of our results and \cite{yang2016estimation}.
10 frames sampled evenly from the first 50 frames of each sequence are used to obtain 10 correspondence sets.
In \figref{fig:Inria_results_pose} we report the average errors for all frames and correspondence sets; our method achieves
state of the art results in pose estimation.
%
%
In the first row of \figref{fig:Dancer_results} we present qualitative results for the INRIA dataset.
Our results are plausible estimates of minimally-clothed shapes.
%
In the second row of \figref{fig:Dancer_results} we qualitatively compare our results to previous work
on the dancer sequence from \cite{de2008performance}. 
Our results visually outperform previous state of the art.
Additional results are presented in the Sup.~Mat; results are best seen in the video included at {\small \url{http://buff.is.tue.mpg.de/}}.

\newcommand{\resinriaourswidth}{0.24\columnwidth}


\begin{figure}
\centering
\includegraphics[width=\resinriaourswidth]{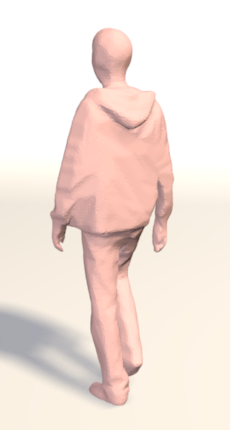} 
\includegraphics[width=\resinriaourswidth]{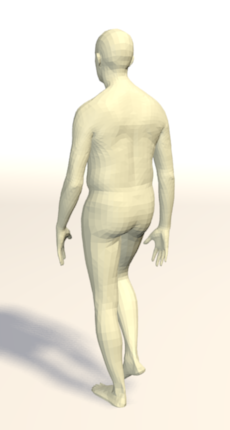} 
\includegraphics[width=\resinriaourswidth]{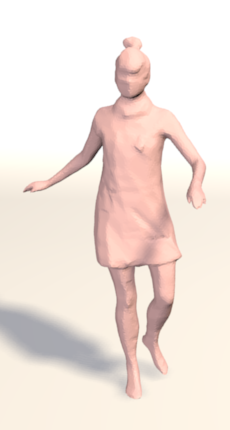} 
\includegraphics[width=\resinriaourswidth]{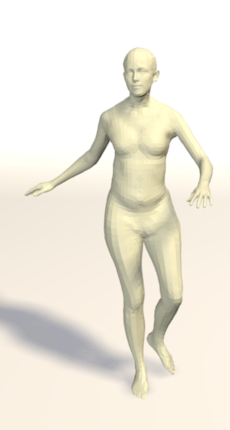}  \\
\includegraphics[width=\resinriaourswidth, trim={270 70 230 250},clip]{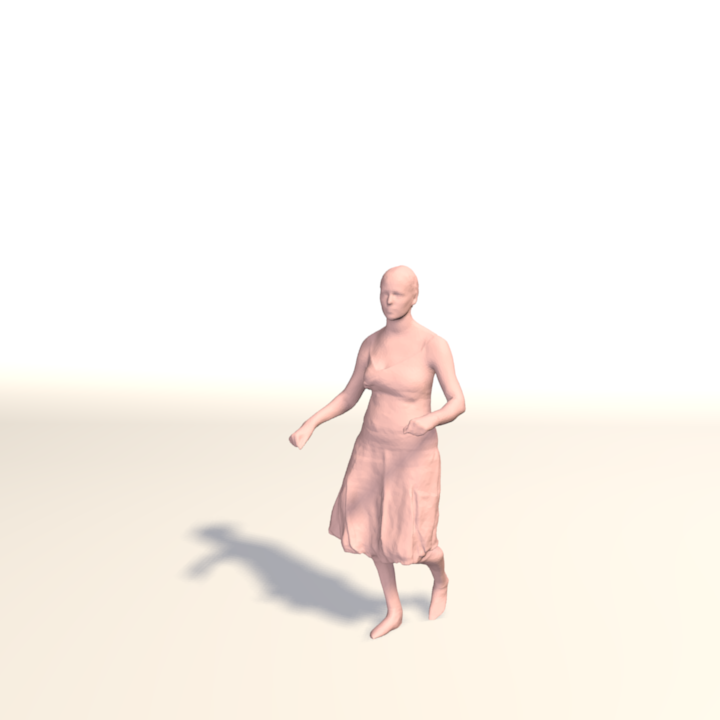} 
\includegraphics[width=\resinriaourswidth, trim={270 70 230 250},clip]{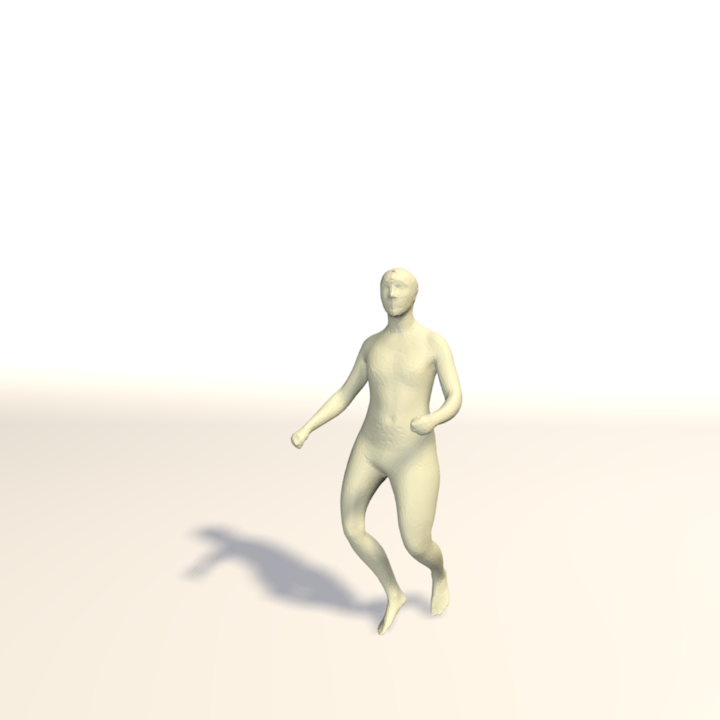} 
\includegraphics[width=\resinriaourswidth, trim={270 70 230 250},clip]{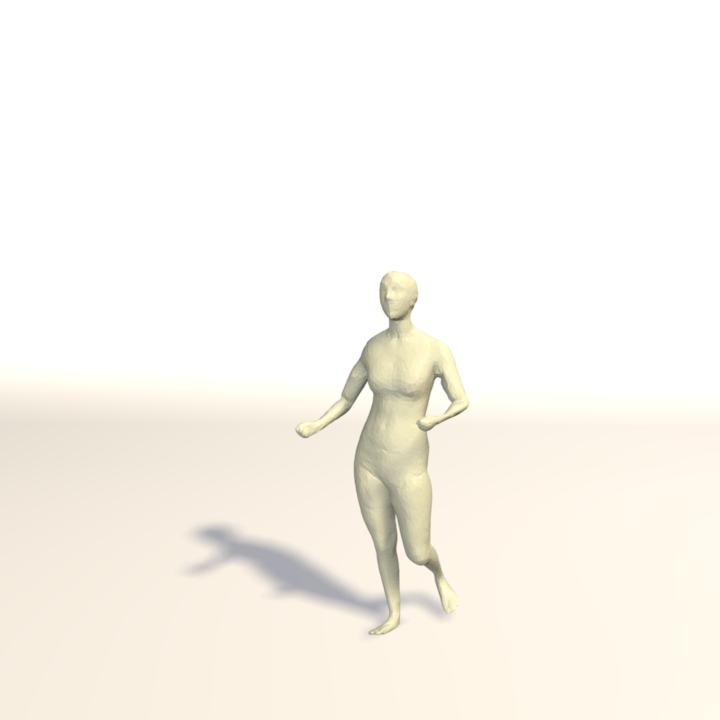} 
\includegraphics[width=\resinriaourswidth, trim={270 70 230 250},clip]{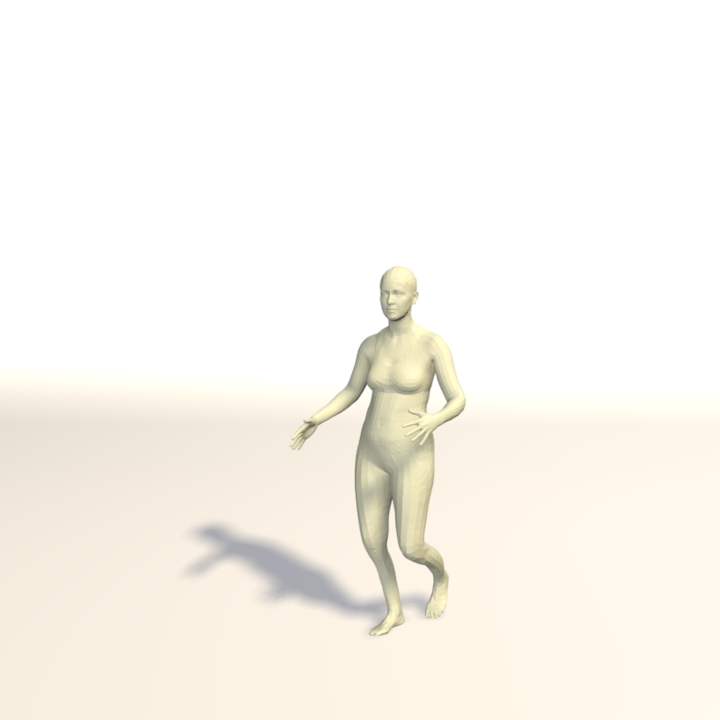}\\
\caption{Top: Qualitative results on the INRIA dataset; scan (pink), our result.
Bottom: Qualitative comparison on Dancer sequence \cite{de2008performance}.
From left to right: scan, Wuhrer \etal~\cite{wuhrer2014estimation}, Yang \etal~\cite{yang2016estimation}, our result.
}
\vspace{-5pt}
\label{fig:Dancer_results} 
\end{figure}

\subsection{Evaluation on \dataset}
To quantitatively evaluate the results in BUFF,
we compare the estimated body shapes with the computed ground truth meshes (\sectref{sec:gt_naked}).
We define the ``registration error'' of the estimated body shape as
the scan-to-model distance with respect to the groundtruth
MCS. 
Given a result mesh $S$, we optimize for pose $\pose$ so that the posed $\vtGT$ best fits $S$.
Then the error between $S$ and the posed $\vtGT$ is computed as the Euclidean distance between
each vertex in $\vtGT$ and its closest point on the surface $S$.

In \tabref{tab:quantitative_results} we show the numerical results obtained by \cite{yang2016estimation},
our fusion mesh, and our detailed mesh.
The results obtained with our method systematically outperform the best state of the art method.
In \figref{fig:NakeCap_results_pose} we show qualitative results on the pose estimations.
Our method properly recovers the scan pose, and visually outperforms \cite{yang2016estimation},
especially in elbow and shoulder estimations.
In \figref{fig:NakeCap_results_shape} we show qualitative results of the shape estimations.
The proposed fusion shape accurately recovers the body shape,
while the detailed shape is capable of capturing the missing details.
While the detailed shape is visually closer to the ground truth,
quantitatively, both results are very similar, see \tabref{tab:quantitative_results}.
In order to evaluate the robustness of the method when skin/cloth segmentation is not available we
evaluate our method labeling the scans of BUFF as \emph{all cloth}. 
While the obtained shapes are less detailed, they are still accurate.
The obtained mean error is $\approx 3$mm (all cloth) compared to $\approx 2.5$mm (detailed) when using our proposed full method.
Additional results and baselines are presented in the Sup.~Mat.


{\bf Computation time and parameters.}
The single-frame objective computation takes $\sim$10 seconds per frame, 
fusion mesh is computed in $\sim$200 seconds.
The detail refinement takes $\sim$40 seconds per frame.
Sequences are computed in parallel and 
computations are executed on an 3GHz 8-Core Intel Xeon E5.
Shapes on BUFF were estimated using $\lambda_\mathrm{skin} = 100$, 
$\lambda_\mathrm{outside} = 100$, $\lambda_\mathrm{fit} = 3$ and
$\lambda_\mathrm{cpl} = 1$.
For INRIA data we decreased the fit term $\lambda_\mathrm{fit} = 1$ to be more
robust to wide clothing. More details in Sup.~Mat.

\section{Conclusion}
\label{sec:conclusion}
We introduced a novel method to estimate a detailed 
body shape under clothing from a sequence of 3D scans.
Our method exploits the information 
in a sequence by fusing all clothed registrations into a single frame.
This results in very accurate shape estimates. 
We also contribute a new dataset (BUFF) of 
high resolution 3D scan sequences of clothed
people as well as ground truth minimally-clothed shapes for each subject.  
\dataset is the first dataset of high quality 4D scans of clothed people; it will 
enable accurate quantitative evaluation of body shape estimation.
Results on \dataset reveal a clear improvement with respect
to the state of the art. 
One of the limitations of the presented approach is the
underestimation of female breast shape; this appears to be a
limitation of SMPL.
SMPL does not take into account soft tissue deformations
of the body; 
future work will incorporate soft tissue deformation models \cite{pons2015dyna}
to obtain even more accurate results.  
In addition, using the
obtained minimally-clothed shapes and cloth alignments we plan to learn a model
of cloth deviations from the body. To collect data to learn such model, we will investigate
the use of Inertial Measurement Units (IMU)~\cite{SIP,vonMarcard:PAMI:2016} to obtain even 
more accurate estimates of pose under wide clothing.

\section{Acknowledgments}
\label{sec:acknowledgments}

We thank the authors of \cite{neophytou2014layered} and
\cite{wuhrer2014estimation} for providing their results for comparison.
We  especially thank the authors of \cite{yang2016estimation} for 
running their method on BUFF.

{\small
\bibliographystyle{ieee}
\bibliography{clothedNakedEst}
}

\end{document}